\documentclass[10pt,twocolumn,letterpaper]{article}

\usepackage{cvpr}     
\usepackage{svg}% To produce the CAMERA-READY version
\usepackage{algorithm}    
\usepackage{algorithmic} 
\usepackage{amsmath}         
\usepackage{amssymb}       
\usepackage{booktabs}       
\usepackage{graphicx}  
\usepackage{multirow}
\usepackage{amssymb}
\usepackage{listings}
\usepackage{xcolor}
\usepackage{cuted}

\usepackage{pifont}% http://ctan.org/pkg/pifont

\definecolor{cvprblue}{rgb}{0.21,0.49,0.74}
\usepackage[pagebackref,breaklinks,colorlinks,allcolors=cvprblue]{hyperref}

\def\paperID{MARS-1} % *** Enter the Paper ID here
\def\confName{TREC VQA}
\def\confYear{2025}

\title{HLTCOE Evaluation Team at TREC 2025: VQA Track}

\author{
Dengjia Zhang$^{1}$\thanks{Equal Contribution.}
\hspace{2.5mm} Charles Weng$^{1}$\footnotemark[1]
\hspace{2.5mm} Katherine Guerrerio$^{1}$ \hspace{2.5mm} Yi Lu$^{1}$ \hspace{2.5mm} \\
Kenton Murray $^{1,2}$ \hspace{2.5mm} Alexander Martin$^{1}$ \hspace{2.5mm} Reno Kriz$^{1,2}$ \hspace{2.5mm} Benjamin Van Durme$^{1,2}$ \hspace{2.5mm} \\
\small ${}^1$Johns Hopkins University \hspace{2.5mm} ${}^2$Human Language Technology Center of Excellence \hspace{2.5mm} \\ 
{\tt\small \{dzhang98, amart233\}@jhu.edu}
}

\begin{document}
\maketitle
% \begin{abstract}
% Video-based question answering (Video QA) remains a core challenge in multimodal understanding due to the complex temporal and causal dynamics of video data. We propose a listwise learning framework that improves semantic precision and ranking consistency in answer generation. Given a video–question pair, a base multimodal model first generates multiple candidate answers, which are then reranked using a model trained with a novel Masked Pointer Cross-Entropy Loss with Rank Weights. This objective integrates pointer-based candidate selection, rank-dependent weighting, and masked cross-entropy under vocabulary restriction, enabling stable and interpretable listwise optimization. By bridging generative modeling with discriminative ranking, our method produces coherent, fine-grained answer lists. Experiments on the TREC2025 VQA dataset show consistent improvements in accuracy and ranking stability, especially for questions requiring temporal reasoning and semantic disambiguation.\footnote{Code: \url{https://github.com/Charles201428/2025-VQA-AG/tree/research-framework-vqa}}
% \end{abstract}

\begin{abstract}
The HLTCOE Evaluation team participated in TREC VQA's Answer Generation (AG) task, for which we developed a listwise learning framework that aims to improve semantic precision and ranking consistency in answer generation. Given a video–question pair, a base multimodal model first generates multiple candidate answers, which are then reranked using a model trained with a novel Masked Pointer Cross-Entropy Loss with Rank Weights. This objective integrates pointer-based candidate selection, rank-dependent weighting, and masked cross-entropy under vocabulary restriction, enabling stable and interpretable listwise optimization. By bridging generative modeling with discriminative ranking, our method produces coherent, fine-grained answer lists. Experiments reveal consistent gains in accuracy and ranking stability, especially for questions requiring temporal reasoning and semantic disambiguation.\footnote{Code: \url{https://github.com/Charles201428/2025-VQA-AG/tree/research-framework-vqa}}
\end{abstract}

%Video-based question answering (Video QA) remains a key challenge in multimodal understanding due to the complex temporal and causal dynamics of video data. We present a listwise learning framework for answer generation and ranking that improves both semantic precision and ordering consistency. Given a video–question pair, a base multimodal model first generates multiple candidate answers, which are then refined by a reranker trained with a novel Masked Pointer Cross-Entropy Loss with Rank Weights. This objective combines pointer-based candidate selection, rank-dependent weighting, and masked cross-entropy under vocabulary restriction, enabling stable and interpretable listwise optimization. The proposed method bridges generative modeling and discriminative ranking, yielding coherent and fine-grained answer rankings. Experiments on TREC2025 VQA show consistent gains in accuracy and ranking stability, particularly on questions requiring temporal reasoning and semantic disambiguation.
% \footnote{* denotes equal contribution in the author list.}    
\section{Introduction}

% The rapid improvement of vision-language models (VLMs) has advanced multimodal understanding across a variety of tasks in text, image, audio, and video~\cite{comanici2025gemini,xu2025qwen2,xu2025qwen3}. Despite this, video understanding remains particularly challenging; unlike static visual scenes, videos require models to integrate spatial, auditory, and temporal cues to construct coherent representations over time~\cite{zohar2025apollo, lin2025unleashinghourscalevideotraining, maaz2023video, martin2025wikivideoarticlegenerationmultiple, shu2025video}. Developing systems capable of such reasoning is a key step toward general multimodal intelligence.
%The rapid advancement of vision-language models has significantly expanded the scope of machine understanding across diverse modalities such as text, image, audio, and video \cite{comanici2025gemini,xu2025qwen2,xu2025qwen3}. Among these, video understanding remains one of the most challenging frontiers due to the inherently dynamic and multi-layered nature of video data. Unlike static visual scenes, videos encode complex temporal and causal relationships, often requiring models to integrate spatial, auditory, and temporal cues to construct coherent representations and perform reasoning over time~\cite{zohar2025apollo, lin2025unleashinghourscalevideotraining, maaz2023video, martin2025wikivideoarticlegenerationmultiple, shu2025video}. Developing systems that can not only perceive but also reason about such dynamic content is thus a crucial step toward general multimodal intelligence.

The TREC VQA answer generation (AG) task requires systems to generate multiple plausible answers per query and return them as a ranked list. Such tasks continue to pose difficulties for current VLMs, which often struggle to generate diverse, well-ranked answer candidates and to maintain stable orderings in open-ended settings~\cite{chen2024measuring,fu2025video,park2025assessing}.
%A fundamental capability within this broader goal is video-based question answering (Video QA)—the task of generating natural language responses to questions grounded in video content. This problem requires aligning visual and textual semantics while maintaining temporal consistency and contextual coherence \cite{zhong2022video,piergiovanni2022video,kim2021video}. In this paper, we focus on the TREC VQA task, which aims to generate multiple answers for each query and provide the generated result as a ranked list of answers. Although recent VLMs \cite[e.g.,][]{xu2025qwen2} have shown impressive progress in this direction, they often struggle with generating diverse answer candidates, ranking multiple plausible answers, modeling fine-grained semantic distinctions, and producing outputs with stable ordering under open-ended settings \cite{chen2024measuring,fu2025video,park2025assessing}. 

To address these limitations, we adopt a listwise learning framework for answer generation and ranking. Our approach is inspired by multi-hypothesis generation methods \cite{vijayakumar2016diverse,fu2023generate,dua2021beyond, martin2025wikivideoarticlegenerationmultiple}, which encourage diverse and expressive candidate answers, as well as listwise reranking techniques \cite{cao2007learning, xia2008listwise, vinyals2015pointer}, which provide principled mechanisms for ordering candidate outputs. Given a video and an associated question, our system first generates multiple candidate answers using a base VLM, and then reorders them using a reranker. For reranking, we introduce a novel \textbf{Masked Pointer Cross-Entropy Loss with Rank Weights}, which integrates three key mechanisms. First, rank-labeled candidate representation explicitly encodes ordinal relations among answer candidates. Next, masked pointer cross-entropy facilitates permutation-consistent optimization through selective supervision of ranked outputs. Finally, rank-dependent weighting enhances the learning signal for higher-priority predictions, while maintaining balanced gradient propagation. Together, these mechanisms allow the model to capture both ranking consistency and linguistic fluency, effectively bridging the gap between generative modeling and discriminative ranking.
%To address these limitations, we explore a listwise learning framework for answer generation and ranking. Specifically, given a video and an associated question, our system generates multiple candidate answers using a base VLM and subsequently refines their ordering through a specialized reranker. The reranker is trained to learn a structured mapping between semantic relevance and answer rank, encouraging consistent and interpretable ranking behavior. To do this reranking, we introduce a novel \textbf{Masked Pointer Cross-Entropy Loss with Rank Weights}, which integrates three key mechanisms: (1) rank-labeled candidate representation, enabling explicit encoding of ordinal relations among answer candidates; (2) masked pointer cross-entropy, facilitating permutation-consistent optimization through selective supervision over ranked outputs; (3) rank-dependent weighting, enhancing the learning signal for higher-priority predictions while maintaining balanced gradient propagation. This formulation allows the model to jointly capture ranking consistency and linguistic fluency, effectively bridging the gap between generative modeling and discriminative ranking.

Our contributions can be summarized as follows:
\begin{enumerate}
    \item We propose a generate-and-rerank setting for producing diverse ranked lists in open-ended Video QA. 
    \item We introduce \textbf{Masked Pointer Cross-Entropy Loss with Rank Weights}, explicitly incorporating ranking information into the optimization process.
    \item We present detailed experiments demonstrating that our proposed pipeline yields consistent improvements over baseline models across multiple evaluation metrics.
\end{enumerate}
\section{Method}

\begin{figure*}[t]
  \centering
  \includegraphics[width=\linewidth]{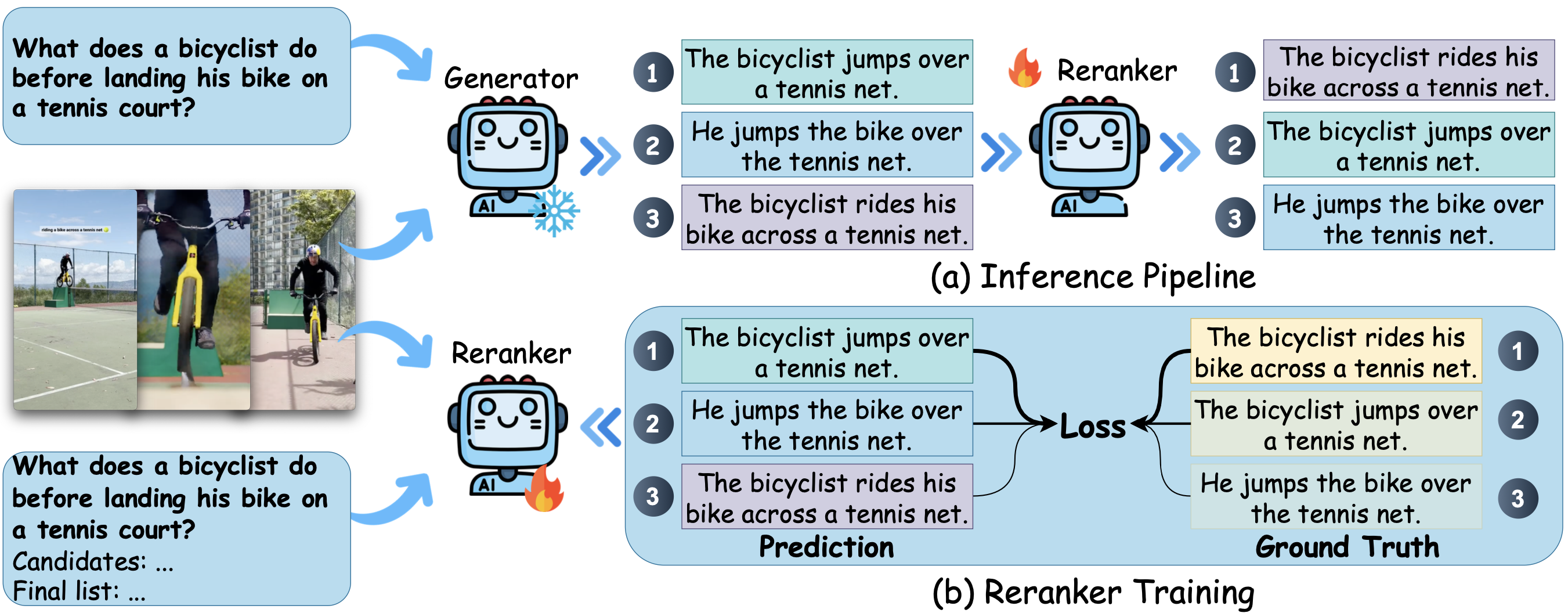}
  % \vspace{-1.5em}
  \caption{Overview of the proposed generate–rank framework. The framework comprises two key components: a Generator, which produces a set of candidate responses given an input prompt, and a Reranker, which assigns relevance scores to these candidates and orders them accordingly.
(a) \textbf{Inference Pipeline}: The generator outputs multiple candidate responses, which are subsequently reordered by the reranker based on their predicted relevance to the query.
(b)\textbf{ Reranker Training}: The reranker is trained using ground-truth rankings, where token-level weighting is incorporated into the loss function to emphasize more informative tokens. Darker colors denote higher token weights, indicating a greater contribution to the overall loss.}
  \label{fig:pipeline}
\end{figure*}

Our approach consists of two main components: 
(1) a \textbf{generator} that produces multiple candidate answers given a query, 
and (2) a \textbf{reranker} that refines these candidates into a ranked list using a masked pointer-based objective.  
An overview of the full pipeline is shown in Figure~\ref{fig:pipeline}.

\subsection{Generator}

The generator module is responsible for producing a diverse set of candidate answers for each input query $q$.  
For each query, we generate $N$ candidates
$\{a_1, a_2, \ldots, a_N\}$, which are subsequently used as input to the reranker.

To encourage diversity, we employ a collection of prompt templates that differ in phrasing, structure, and reasoning cues.  
Instead of tuning model parameters or sampling hyperparameters, we rely on prompt variation as a lightweight and reproducible way to obtain diverse candidates.  
These prompts are empirically evaluated on a validation set, 
and the template that yields the best reranking performance is selected and fixed for both training and evaluation.  
This setup ensures that the generation process remains deterministic and reproducible, 
while the resulting candidate pool maintains sufficient diversity.  

\subsection{Reranker}

The reranker aims to reorder the generated candidates 
$\{a_1, a_2, \ldots, a_N\}$ into a coherent ranked list that aligns with ground-truth relevance.  
It is trained using a \textbf{Masked Pointer Cross-Entropy Loss with Rank Weights}, 
which integrates three key mechanisms:
(1) Rank-Labeled Candidate Representation, 
(2) Masked Pointer Cross-Entropy, and 
(3) Rank-Dependent Weighting.

\subsubsection{Rank-Labeled Candidate Representation}

For each query $q$ with its reference answer $a^{*}$, 
we first compute the BERTScore \cite{zhang2019bertscore} between each generated candidate $a_i$ and $a^{*}$.  
Candidates are then ordered to form a ranked sequence in a fixed way:
\begin{verbatim}
    <R1> = <CAND_...>
    <R2> = <CAND_...>
    ...
    <RN> = <CAND_...>
    <ENDLIST>
\end{verbatim}

\noindent Each token in this sequence represents either a rank marker (\texttt{<Rn>}), 
a candidate pointer (\texttt{<CAND\_m>}), 
or regular textual content.  
This rank-labeled sequence serves as the target supervision for the reranker.

\subsubsection{Masked Pointer Cross-Entropy}

The reranker is trained to autoregressively generate the rank-labeled sequence.  
At each decoding step $t$, the model produces logits $\mathbf{z}_t \in \mathbb{R}^{V}$ over the vocabulary.  
When the target token corresponds to a candidate pointer (e.g., \texttt{<CAND\_3>} 
or the termination token \texttt{<ENDLIST>}), 
the prediction space is restricted to the subset of valid, unselected candidates $\mathcal{C}_k$:
\begin{equation}
P(y^{(k)}|\mathcal{C}_k) = 
\frac{\exp(z^{(k)}_{y^{(k)}})}{\sum_{j \in \mathcal{C}_k} \exp(z^{(k)}_j)}.
\end{equation}
The corresponding pointer loss is:
\begin{equation}
\mathcal{L}^{(k)}_{\text{ptr}} = - \log P(y^{(k)} | \mathcal{C}_k).
\end{equation}

For non-pointer tokens (e.g., rank markers or normal text), 
the model computes standard cross-entropy over the full vocabulary:
\begin{equation}
\mathcal{L}^{(t)}_{\text{text}} = - \log 
\frac{\exp(z^{(t)}_{y_t})}{\sum_{v=1}^{V} \exp(z^{(t)}_v)}.
\end{equation}

\begin{table*}[!htbp]
\centering
\setlength{\tabcolsep}{4.5pt}
\begin{tabular}{llcc|cc|cc}
\toprule
\multirow{2}{*}{\textbf{Split}} &
\multirow{2}{*}{\textbf{Settings}} 
& \multicolumn{2}{c|}{\textbf{nDCG}} 
& \multicolumn{2}{c|}{\textbf{BERTScore}} 
& \multicolumn{2}{c}{\textbf{METEOR}} \\
\cmidrule(r){3-4} \cmidrule(r){5-6} \cmidrule(r){7-8}
& & \textbf{BERTScore Rank} & \textbf{METEOR Rank} 
  & \textbf{Mean} & \textbf{Max} 
  & \textbf{Mean} & \textbf{Max} \\
\midrule
\multirow{3}{*}{Dev Data} 
& QwenOmni \textbf{Generator} & 0.997 & \textbf{0.772} & 0.869 & 0.875 & 0.145 & 0.182 \\
& QwenVL \textbf{Generator}             & 0.995 & 0.664 & \textbf{0.904} & 0.905 & \textbf{0.175} & 0.183 \\
& QwenVL \textbf{Generator}+\textbf{Reranker} & \textbf{0.998} & 0.742 & \textbf{0.904} & \textbf{0.916} & \textbf{0.175} & \textbf{0.250} \\

\midrule
\multirow{2}{*}{Test Data} 
&QwenVL \textbf{Generator} &  0.995 & 0.708 & \textbf{0.896} & 0.896 & \textbf{0.157} & 0.159 \\
&QwenVL \textbf{Generator}+\textbf{Reranker} & \textbf{0.995} & \textbf{0.718} & \textbf{0.896} & \textbf{0.897} & 0.156 & \textbf{0.161} \\
\bottomrule
\end{tabular}
\caption{Comprehensive evaluation of different generator configurations on the TRECVID AG task. 
Results are reported on both the development and test splits under three metrics: NDCG, BERTScore, and METEOR. 
QwenVL achieves superior overall performance compared to QwenOmni, particularly in semantic similarity metrics (BERTScore and METEOR). 
The inclusion of a reranker further refines ranking quality, yielding the most notable gains under the METEOR metric.
}
\label{tab:main}
% \vspace{-1.5em}
\end{table*}

\subsubsection{Rank-Dependent Weighting}

To emphasize higher-ranked predictions, 
we assign a weight $w_k$ to each pointer step corresponding to rank $k$:
\begin{equation}
w_k =
\begin{cases}
\alpha_k, & \text{if } k \leq 3, \\
1, & \text{otherwise},
\end{cases}
\end{equation}
where $\alpha_k > 1$ are hyperparameters controlling the relative importance of top-ranked predictions.
These weights are applied directly to the loss terms at respective rank steps.

\subsubsection{Final Objective}

The final training objective aggregates all weighted losses:
\begin{equation}
\mathcal{L}_{\text{total}} =
\frac{1}{\sum_t w_t} 
\sum_t w_t \, \mathcal{L}_t,
\end{equation}
where $\mathcal{L}_t$ is either $\mathcal{L}^{(k)}_{\text{ptr}}$ or $\mathcal{L}^{(t)}_{\text{text}}$. This formulation ensures:
\begin{enumerate}
    \item candidate tokens are compared only within the subset of valid remaining candidates;
    \item higher-ranked predictions receive stronger supervision;
    \item gradient flow remains stable without in-place masking.
\end{enumerate}

The proposed \textbf{Masked Pointer Cross-Entropy with Rank Weights} 
thus enables the model to learn fine-grained correspondences between candidate order and predicted likelihood, 
achieving stable and consistent listwise reranking performance.

\section{Experiment}
% \subsection{Settings}
\paragraph{Data and Models} We designate the first 30 examples of the official training data provided for the TREC VQA AG task as an ad-hoc dev split and use the remainder as our actual train split. We report results both on this ad-hoc dev split and on the official test split. For the \textbf{generator}, we evaluate both Qwen2.5-VL-7B~\cite{bai2025qwen25vltechnicalreport} and Qwen2.5-Omni-7B~\cite{xu2025qwen25omnitechnicalreport} to compare their generative capabilities. For the \textbf{reranker}, we employ Qwen2.5-VL-7B as the backbone model and fine-tune it specifically for ranking refinement.

% \subsection{Evaluation Metrics}
% To assess the quality of ranked answers, we adopt the same metrics as the TREC shared task. Specifically, we report normalized discounted cumulative gain (nDCG) for two metrics, BERTScore \cite{zhang2019bertscore} and METEOR \cite{banerjee2005meteor}. We also report both the mean, the average score across all generated answers, and the best, the score corresponding to the top-ranked (first) answer, for BERTScore and METEOR. For our ablations on prompting, we also report ROUGE-L \cite{lin-2004-rouge} and Semantic Text Similarity \cite[STS;][]{cer-etal-2017-semeval}.

% To rigorously assess the quality of ranked answers, we adopt the \textbf{Normalized Discounted Cumulative Gain (nDCG)} metric under two relevance functions: \textbf{BERTScore} and \textbf{METEOR}.

% \begin{itemize}
%     \item \textbf{NDCG-BERTScore} and \textbf{NDCG-METEOR} evaluate ranking quality when the relevance between a generated answer and the reference is measured using BERTScore and METEOR, respectively.
%     \item For other metrics, we report two variants:
%     \begin{itemize}
%         \item \textbf{Mean}: the average score across all generated answers.
%         \item \textbf{Best}: the score corresponding to the top-ranked (first) answer.
%     \end{itemize}
% \end{itemize}

\subsection{Results}
The overall results are presented in Table~\ref{tab:main}.
We evaluate two main configurations:
\begin{enumerate}
    \item \textbf{Generator}: a single-stage model that directly generates candidate answers.
    \item \textbf{Generator + Reranker}: a two-stage framework where the generator produces candidate answers and the reranker refines their ranking.
\end{enumerate}

For the dev split, we additionally test the Qwen Omni Generator to investigate the performance of the Omni variant.
Although Qwen Omni achieves slightly higher NDCG scores, it performs worse in BERTScore and METEOR.
Since our goal is to achieve a balanced improvement across all evaluation metrics, and as shown in \autoref{prompt_study}, fine-tuning the generator alone does not substantially enhance semantic alignment, \textit{we ultimately adopt Qwen-VL as the generator in our final system.}

\begin{table}[]
\centering
\begin{tabular}{lccc}
\toprule
Prompt & ROUGE-L & METEOR  & STS \\
\midrule
PROMPT1 & 0.224 & 0.222  & 0.469 \\
PROMPT2 & 0.334 & 0.216  & 0.607 \\
PROMPT3 & 0.389 & 0.228  & 0.631 \\
PROMPT4 & 0.383 & 0.227  & 0.623 \\
PROMPT5 & \textbf{0.407} & \textbf{0.251}  & \textbf{0.638} \\
PROMPT6 & 0.390 & \underline{0.246}  & 0.620 \\
PROMPT7 & 0.352 & 0.245 & 0.595 \\
% PROMPT5 w/ lora & 0.4014 & 0.2402 & 0.6298 \\
% PROMPT6 w/ lora & 0.3797 & 0.2406 & 0.6220 \\
PROMPT5 w/ lora & \underline{0.394} & 0.239 & \underline{0.636} \\
\bottomrule
\end{tabular}
\caption{Ablation results of different prompt templates and their LoRA variants on our ad-hoc dev split.}
\label{tab:prompt_raw}
\end{table}
\subsection{Prompt Study} \label{prompt_study}
To further examine the influence of instruction formulation and fine-tuning strategies, we conduct an ablation study on prompt templates and LoRA-based fine-tuning.\footnote{See \autoref{appendix:parameters} for the parameters used in training. The prompt variants can be found in \autoref{sup:prompts}.}
Each prompt variant represents a distinct instruction template used during generation.
% The suffix \textit{lora} denotes that the generator has been fine-tuned using LoRA with the respective prompt.

Table~\ref{tab:prompt_raw} reports ROUGE-L \cite{lin-2004-rouge}, METEOR \cite{banerjee-lavie-2005-meteor}, and STS on our ad-hoc dev split.
Among all prompt variants, Prompt5 consistently achieves the best overall performance across metrics.
\textit{Accordingly, we adopt Prompt5 as the final instruction template for the generator in our final system.}

% \begin{table}[ht]
% \centering
% \begin{tabular}{lcccc}
% \toprule
% Prompt & ROUGE\_L & METEOR & BERTScore & STS \\
% \midrule
% PROMPT1 & 0.2243 & 0.2227 & 0.2515 & 0.4691 \\
% PROMPT2 & 0.3342 & 0.2164 & 0.4762 & 0.6079 \\
% PROMPT3 & 0.3894 & 0.2281 & 0.4970 & 0.6314 \\
% PROMPT4 & 0.3837 & 0.2278 & 0.4849 & 0.6235 \\
% PROMPT5 & 0.4070 & 0.2518 & 0.5081 & 0.6380 \\
% PROMPT6 & 0.3905 & 0.2461 & 0.4849 & 0.6209 \\
% PROMPT7 & 0.3521 & 0.2456 & 0.4416 & 0.5952 \\
% PROMPT5\_raw\_lora & 0.4014 & 0.2402 & 0.4963 & 0.6298 \\
% PROMPT6\_raw\_lora & 0.3797 & 0.2406 & 0.4922 & 0.6220 \\
% PROMPT5\_prompt\_lora & 0.3949 & 0.2397 & 0.5019 & 0.6368 \\
% \bottomrule
% \end{tabular}
% \caption{Raw evaluation scores of different prompts and LoRA variants.}
% \label{tab:prompt_raw}
% \end{table}

\section{Conclusion/Discussion}
Our experiments have aimed to assess the effectiveness of our proposed framework on the TREC VQA AG task, examining the influence of prompt design on the generator’s performance and the usefulness of weighted reranking for optimizing the ordering of candidate answers. Results demonstrate that well-crafted prompts play a crucial role in producing semantically accurate and contextually relevant responses. 
Moreover, the proposed \textbf{Masked Pointer Cross-Entropy Loss with Rank Weights} substantially enhances ranking consistency and improves the quality of top-ranked answers. 
We believe that this training paradigm provides valuable insights for future research on rank-list optimization and could serve as a general framework for learning-to-rank in multimodal generation tasks.

% \section*{Limitation and Future Work }
% Given that the current framework is trained exclusively on the TREC dataset, its overall performance may be constrained by the limited diversity and scale of the training data. Enhancing the reranker through exposure to additional datasets could therefore lead to more robust and generalizable results. Similarly, the suboptimal performance of the generator may also be attributed to the restricted amount of training data. Jointly training the generator and reranker on a larger and more diverse corpus could further improve the quality and consistency of generated outputs.
% Furthermore, the proposed framework can be extended to other promising research directions, such as reinforcement learning–based reasoning mechanisms for QA tasks. For instance, within the GRPO framework \cite{shao2024deepseekmath}, the reranker can be incorporated into the rollout phase to assess and rank the generated candidate answers, thereby providing differentiated auxiliary rewards according to their relative quality, rather than depending solely on a verification-based reward signal. This integration could enable a more refined and effective optimization process.
%\input{sec/1_intro}
% \input{sec/1_intro}
% \input{sec/2_formatting}
% \input{sec/3_finalcopy}
{
    \small
    \bibliographystyle{ieeenat_fullname}
    \bibliography{main}
}
\clearpage

% \clearpage

% \setcounter{page}{1}
% \maketitlesupplementary
% \begin{strip}

\appendix

\section{Model Parameters}
\label{appendix:parameters}

For Generator, we use Qwen2.5-VL 7B as the generator model, with the decoding parameters set to \texttt{temperature = 0.7} and \texttt{top\_p = 0.9}.

For Reranker Training, we adopt LoRA Fine-tuning. The training dataset is from TREC VQA AG dataset, containing a total 500 samples and we choose 90\% of them for training and rest of them for evaluating. The reranker’s weights are set as \( w_1 = 3 \), \( w_2 = 1.5 \), and \( w_3 = 1.2 \). 
The training hyperparameters are as follows: 
\texttt{learning rate = 2e-6}, \texttt{batch size = 1}, and \texttt{epochs = 5}.

For the LoRA configuration, we set \texttt{lora\_r = 16}, 
\texttt{lora\_alpha = 32}, and \texttt{lora\_dropout = 0.05}.

\section{Prompts}\label{sup:prompts}
In this section we provide the prompt variations used in \autoref{prompt_study} for Prompts 1-7 in \autoref{prompt:1}, \autoref{prompt:2}, \autoref{prompt:3}, \autoref{prompt:4}, \autoref{prompt:5}, \autoref{prompt:6}, and \autoref{prompt:7}, respectively.

\begin{figure*}[t]
\noindent\fbox{%
    \parbox{.98\textwidth}{%
\footnotesize
{\tt
\small
Answer the following question concisely in one sentence:

question:
{question}
}
}}
\caption{{Prompt1}}
\label{prompt:1}
\end{figure*}

\begin{figure*}[t]
\noindent\fbox{%
    \parbox{.98\textwidth}{%
\footnotesize
{\tt
\small
Answer the following question concisely in one sentence, you should follow the points:

1. You should answer the question as simple as possible, some questions may just need a word or two.\\
2. You don't need to answer the question in a very detailed way, just give a concise answer.\\

question:
{question}
}
}}
\caption{{Prompt2}}
\label{prompt:2}
\end{figure*}

\begin{figure*}[t]
\noindent\fbox{%
    \parbox{.98\textwidth}{%
\footnotesize
{\tt
\small
Answer the following question concisely in one sentence, you should follow the points:

1. You should answer the question as simple as possible, some questions may just need a word or two.\\
2. You don't need to answer the question in a very detailed way, just give a concise answer.\\
3. For the answer in number, you should answer the number (one, two, three, etc.) but not 1,2,3, etc. in the question.\\

question:
{question}
}
}}
\caption{{Prompt3}}
\label{prompt:3}
\end{figure*}

\begin{figure*}[t]
\noindent\fbox{%
    \parbox{.98\textwidth}{%
\footnotesize
{\tt
\small
Answer the following question concisely in one sentence, you should follow the points:

1. You should answer the question as simple as possible, some questions may just need a word or two.\\
2. You don't need to answer the question in a very detailed way, just give a concise answer.\\
3. For the answer in number, you should answer the number (one, two, three, etc.) but not 1,2,3, etc. in the question.\\
4. You need to think about the question and answer it carefully.\\

question:
{question}
}
}}
\caption{{Prompt4}}
\label{prompt:4}
\end{figure*}

\begin{figure*}[t]
\noindent\fbox{%
    \parbox{.98\textwidth}{%
\footnotesize
{\tt
\small
Answer the following question concisely in one sentence, you should follow the points:

1. You should answer the question as simple as possible, some questions may just need a word or two.\\
2. You don't need to answer the question in a very detailed way, just give a concise answer.\\
3. For the answer in number, you should answer the number (one, two, three, etc.) but not 1,2,3, etc. in the question.\\

the transcript of the video is:
{transcript}

question:
{question}
}
}}
\caption{{Prompt5}}
\label{prompt:5}
\end{figure*}

\begin{figure*}[t]
\noindent\fbox{%
    \parbox{.98\textwidth}{%
\footnotesize
{\tt
\small
Answer the following question concisely in one sentence, you should follow the points:

1. You should answer the question as simple as possible, some questions may just need a word or two.\\
2. You don't need to answer the question in a very detailed way, just give a concise answer.\\
3. For the answer in number, you should answer the number (one, two, three, etc.) but not 1,2,3, etc. in the question.\\
4. You need to think about the question and answer it carefully.\\
5. If the question is about the audio, you should answer the question based on the transcript of the video.\\
6. The transcript can be in different languages, you should answer the question in English.\\

the transcript of the video is:
{transcript}

question:
{question}
}
}}
\caption{{Prompt6}}
\label{prompt:6}
\end{figure*}

\begin{figure*}[t]
\noindent\fbox{%
    \parbox{.98\textwidth}{%
\footnotesize
{\tt
\small
Answer the following question concisely in one sentence, you should follow the points:

1. You should answer the question as simple as possible, some questions may just need a word or two.\\
2. You don't need to answer the question in a very detailed way, just give a concise answer.\\
3. For the answer in number, you should answer the number (one, two, three, etc.) but not 1,2,3, etc. in the question.\\
4. The ASR transcript can be in different languages, you should answer the question in English.\\
5. Base your answer on both the ASR transcript and the video frames; if needed, use reasoning or general knowledge to give the most reasonable answer.\\

the ASR transcript of the video is:
{transcript}

the question is:
{question}
}
}}
\caption{{Prompt7}}
\label{prompt:7}
\end{figure*}

\end{document}